\documentclass[runningheads]{llncs}
\usepackage{graphicx}
\usepackage{amsmath,amssymb,amsfonts}
\usepackage{algorithmic}
\usepackage{textcomp}
\usepackage{xcolor}
\usepackage{url}
\usepackage{float}
\usepackage{commath}
\usepackage[caption=false]{subfig}
\usepackage{appendix}
\usepackage{tabularx}
\usepackage{multirow}
\begin{document}
\title{Deep-Motion-Net: GNN-based volumetric organ shape reconstruction from single-view 2D projections}
\titlerunning{Deep-Motion-Net}
%
\author{Isuru Wijesinghe\inst{1,2} \and
Michael Nix\inst{3} \and
Arezoo Zakeri\inst{4} \and
Alireza Hokmabadi\inst{5} \and
Bashar Al-Qaisieh\inst{3}
Ali Gooya\inst{6} \and
Zeike A. Taylor\inst{1,2}}
\authorrunning{I. Wijesinghe et al.}
%
\institute{Centre for Computational Imaging and Simulation Technologies in Biomedicine, School of Mechanical Engineering, University of Leeds, Leeds, UK 
\and 
Institute of Medical and Biological Engineering, School of Mechanical Engineering, University of Leeds, Leeds, UK 
\and
Department of Medical Physics and Engineering, St James’s University Hospital, Leeds Teaching Hospitals NHS Trust, Leeds, UK
\and
Division of Informatics, Imaging, and Data Sciences, Faculty of Biology, Medicine and Health, University of Manchester, Manchester, UK
\and
School of Medicine and Population Health, University of Sheffield, Sheffield, UK
\and
School of Computing Science, University of Glasgow, Glasgow, UK \\
}
\maketitle              
\begin{abstract}
We propose Deep-Motion-Net: an end-to-end graph neural network (GNN) architecture that enables 3D (volumetric) organ shape reconstruction from a single in-treatment kV planar X-ray image acquired at any arbitrary projection angle. Estimating and compensating for true anatomical motion during radiotherapy is essential for improving the delivery of planned radiation dose to target volumes while sparing organs-at-risk, and thereby improving the therapeutic ratio. Achieving this using only limited imaging available during irradiation and without the use of surrogate signals or invasive fiducial markers is attractive. The proposed model learns the mesh regression from a patient-specific template and deep features extracted from kV images at arbitrary projection angles. A 2D-CNN encoder extracts image features, and four feature pooling networks fuse these features to the 3D template organ mesh. A ResNet-based graph attention network then deforms the feature-encoded mesh. The model is trained using synthetically generated organ motion instances and corresponding kV images. The latter is generated by deforming a reference CT volume aligned with the template mesh, creating digitally reconstructed radiographs (DRRs) at required projection angles, and DRR-to-kV style transferring with a conditional CycleGAN model. The overall framework was tested quantitatively on synthetic respiratory motion scenarios and qualitatively on in-treatment images acquired over full scan series for liver cancer patients. Overall mean prediction errors for synthetic motion test datasets were 0.16$\pm$0.13 mm, 0.18$\pm$0.19 mm, 0.22$\pm$0.34 mm, and 0.12$\pm$0.11 mm. Mean peak prediction errors were 1.39 mm, 1.99 mm, 3.29 mm, and 1.16 mm.

\end{abstract}
\section{Introduction}
In this paper, we present a method, labelled Deep-Motion-Net, for estimating 3D volumetric organ deformation from a single in-treatment kV planar X-ray image acquired at any arbitrary gantry (projection) angle. Internal anatomical motion confounds the precise delivery of radiation to target volumes during external beam radiotherapy. Precision is critical for achieving tumour coverage while preserving surrounding sensitive healthy tissues(~\cite{keall2006management}) since normal tissue damage prevents dose escalation to the gross target volume (GTV) to the desired therapeutic level. In hypo-fractionated treatments (i.e. Stereotactic Ablative Body Radiotherapy), wherein higher, but more precisely targeted doses are used, unaccounted motion is yet more critical and sometimes prohibitive. Hence, to fully exploit the potential of external beam radiation, tumour and organ movements must be addressed during irradiation such that more radiation is delivered to the target tumour while sparing organs-at-risk (OARs).

Typically, a cone beam CT image is acquired at the start of treatment and used to ensure the patient is correctly positioned with respect to the linear accelerator (linac). However, this static single-timepoint representation ignores anatomical motion that subsequently occurs during radiation delivery, which may cause overdosing of OARs, or underdosing of tumour, leading to poorer outcomes for survival and post-treatment morbidity (\cite{keall2006management}). To deal with respiratory-induced tumour/organ motions, numerous mitigation strategies have been developed, broadly categorised as either passive or active (\cite{keall2006management,chi2014potential}). Defining an internal-target-volume (ITV) that includes a motion-encompassing margin is an example of a passive mitigation technique(~\cite{chi2014potential}). The motion-encompassing margin is normally estimated based on 4D-CT data acquired at treatment planning and hence reflects an average estimate of the motion that corresponds with in-treatment motion with uncertain accuracy. While these margins allow for inaccuracies during treatment, they also result in greater irradiation of normal tissues(~\cite{gargett2019clinical}), meaning overall radiation intensity must be reduced, and treating the target tumour becomes more difficult.

To minimise, or even eliminate the ITV, effective active motion management is critical. Respiratory-gating is a popular method since it is relatively easy to implement, but it implies a longer time to deliver the specified dose because radiation is only delivered for a segment of the respiratory cycle(~\cite{keall2006management}). In contrast, real-time tracking, which repositions and/or reshapes the radiation beam as the target moves, implies no prolongation of treatment sessions but is more technically challenging to realise. Moreover, its effectiveness can be limited by the time delay between detecting a change in target position and the system adjustment, resulting in a persistent lag in the system's response to the target position.

All such active methods critically rely on real-time information on the tumour position during treatment. Gating and tracking techniques often rely on implanted markers to track the target mobility in real-time. These markers are invasive, and in any case, only provide information on specific locations (i.e., marker positions) inside tissues, rather than the target/OARs as a whole(~\cite{abbas2014motion}). Techniques based on non-invasive imaging are preferred. Treatment systems integrating magnetic resonance imaging (MR-linac) arguably provide an excellent basis for this(~\cite{witt2020mri}) in the form of real-time in-treatment images that are radiation-free and have good soft tissue contrast and resolution(~\cite{paganelli2018mri}). However, current MR-linacs provide only orthogonal pairs of 2D slices rather than true 3D images and hence do not directly enable visualisation of the whole 3D geometry of a target tumour region and surrounding OARs. More importantly, such systems are expensive and rare, meaning very few patients currently can access them.

In contrast, most conventional linacs are equipped with on-board kV X-ray imaging, and such systems will inevitably be used to treat most patients; techniques that can recover anatomical motion from such images are therefore attractive. The key challenges, however, are poor soft tissue contrast and compression of volumetric information along projection directions. Reconstructing high-quality 3D organ shapes from such images is, therefore, difficult. We hypothesise nonetheless that when combined with suitable prior information in the form of learned models of motion patterns and corresponding image appearance, such images do provide sufficient information to recover accurate 3D anatomical information. 

We propose a deep graph neural network (GNN) model for this purpose. The model learns a mapping from kV image-derived features to displacements of nodes in a patient-specific template organ mesh. Features are extracted by a convolutional neural network (CNN) image encoder, while regression of the features with node displacements is learned by a graph attention (GAT) network. Importantly, the complete model is end-to-end trainable by virtue of a series of feature pooling networks (FPNs) that fuse image features with the 3D graph nodes, eliminating non-trainable components (i.e. vertex projection onto the 2D image space) that would otherwise be required. Finally, the model also learns projection angle-dependent features by encoding the angle in an additional channel to the input image. By this means, the model can reconstruct the 3D anatomy from kV images acquired at any projection angle. To the best of our knowledge, this is the first framework capable of reconstructing 3D anatomy from such inputs. While the method is general, we focus in this work on respiratory motion, and evaluate the method using synthetic and real images from liver cancer patients.

\section{Previous work: 3D shape reconstruction from single-view projections}

We here review existing techniques for recovering 3D geometry from 2D images, comparable to our approach. Another major class of techniques are so-called surrogate-driven models, which estimate internal anatomical motion using some surrogate signal under the assumption the two are well correlated. For further information on such approaches, interested readers are referred to \cite{mcclelland2013respiratory}.

\subsection{RGB image-based reconstruction}

Several reports have described techniques for reconstruction from RGB images. For example, ~\cite{wang2018pixel2mesh} proposed the GNN-based Pixel2Mesh algorithm, which deforms an ellipsoidal surface mesh using CNN-derived semantic characteristics from an input image, and applies it to the analysis of natural shapes (aeroplanes, chairs, cars, etc.). The ellipsoidal starting mesh limits the approach to genus-0 shapes, though in principle it could be adapted to other topologies. ~\cite{smith2019geometrics} extended the method to better capture local surface geometry, though the topological constraints remained. Similar ideas were used in \cite{kolotouros2019convolutional,Zhang_2021} to reconstruct 3D human body shapes from single RGB images. In the medical domain, ~\cite{Wu_2019} proposed a CNN architecture for reconstructing 3D lung shapes, in the form of point clouds, from a single-view 2D laparoscopic image.

\subsection{Projection image-based reconstruction}

While clearly sharing elements of our target problem, reconstruction from RGB images rather than 2D projections is nonetheless a substantially different one. Various approaches addressing the latter scenario have appeared recently. \cite{X2CT_GAN} proposed X2CT-GAN to reconstruct 3D-CT volumes from bi-planar 2D X-ray images using generative adversarial networks. Reconstruction of full CT volumes is impressive, however, in our scenario only single-view projections are available, making the problem significantly more challenging.

\cite{Wang_2019} proposed a CNN-based approach for reconstructing lung surface shapes from single-view 2D projections. Their method employs free-form deformation to learn the optimal smooth deformation from multiple templates to match the query 2D image. ~\cite{x_ray2shape} proposed X-ray2Shape to reconstruct 3D liver surface meshes by combining GNN and CNN networks (the latter to encode image features). A mean shape derived from 124 patients was used as prior (i.e. initial template) and deformed by the GNN to match the individual organ shape. Later, the same authors~\cite{nakao2021image,Nakao_9844010} extended the approach to reconstruct multiple abdominal organ shapes from a single projection image. These approaches \cite{Wang_2019,x_ray2shape,nakao2021image,Nakao_9844010} are designed to operate on images acquired at fixed projection angles---front view projections, equivalent to gantry angle 0 in our case---and therefore cannot directly accommodate images from arbitrary angles, as required here.

This limitation was addressed by \cite{wei2020real}, who proposed a CNN architecture to predict principal component analysis (PCA) coefficients in a 4D-CT-based breathing motion model. The authors used an angle-dependent region of interest (ROI) 2D projection mask to remove pixels unrelated to respiration and an angle-dependent fully-connected (FC) layer. This layer was designed to handle discrete (integer-valued) angles ranging from 0 to 360, and only a single group of weights and biases were used to generate its output for each degree of the projection angle. For real, continuously varying angles, binary projection masks at the nearest integer were chosen. However, challenges appeared in cases with significant intensity variations between DRRs and CBCT projections, affecting localization accuracy due to intensity shift issues. Additionally, reconstruction artefacts in 4D-CT images, such as structural blurriness or duplications, influenced localization accuracy. The study suggests potential limitations in handling variations in breathing amplitude and patient setup during treatment, recommending retraining PCA and CNN models with repeated 4D-CT data acquisition for validation. Furthermore, tumour localization using binary projection masks proved challenging for certain cases and projection angles, when the corresponding images contained little information related to breathing motion \cite{wei2020real}.

A further limitation of the foregoing approaches is their prediction only of organ surface shapes, rather than their full volumes. A true volumetric approach for estimating liver deformations was recently proposed by ~\cite{shao2022real}. Their method first used a GNN to predict the deformed liver surface. These deformations were then passed as boundary conditions to a finite element model of the liver, which computed the corresponding volumetric deformations. In this way, some level of biomechanical constraint was also introduced. The approach was evaluated for several projection angles (specifically: $0^\circ$, $45^\circ$, and $90^\circ$). However, the model required retraining for each angle; that is, each new angle effectively required a separate model. The model also required a very high number (3840) of image features to be encoded on graph nodes; in our approach, we use only 20. Finally, while biomechanical constraints can, in principle, be attractive for enforcing physical plausibility, the finite element solutions were, in practice time consuming, which may be significant for clinical use, especially in-treatment adaption of therapy.

All of these 3D geometry reconstructions using single-view projection-based approaches utilized DRR images with a wide field-of-view (FOV) to encompass the entire liver anatomy for their experiments. However, in clinical settings, projection images are usually acquired with a limited FOV. This discrepancy poses challenges for prior methods \cite{Wang_2019,x_ray2shape,nakao2021image,Nakao_9844010,shao2022real} when extracting features by projecting vertices onto the DRR/kV plane, as the projected mesh nodes may extend beyond the boundaries of the projection FOV, thereby hindering effective feature extraction \cite{shao2022real}.

\subsection{Recapitulation}

Several approaches to reconstructing organ geometry from single-view projection images have been proposed. Their main shortcomings, addressed by our proposed method, are as follows:

\begin{itemize}
    \item \emph{Fixed projection angle}: Most existing approaches \cite{Wang_2019,x_ray2shape,nakao2021image,Nakao_9844010,shao2022real} are designed to operate on images acquired at fixed projection angles rather than the varying angles used during RT.
    \item \emph{Surface reconstruction only}: Most existing approaches \cite{Wang_2019,x_ray2shape,nakao2021image,Nakao_9844010,wei2020real} reconstruct only surface representations of the involved organs and do not, therefore, describe deformations within those organs.
    \item \emph{Wide field of view}: Most existing approaches \cite{Wang_2019,x_ray2shape,nakao2021image,Nakao_9844010,shao2022real} depend on DRR images with a wide FOV, not reflective of the usual clinical situation.
\end{itemize}

\section{Methodology}
Building on several of the works cited above, the core of our approach is a GNN that learns mappings from kV image features to nodal displacements of a patient-specific organ mesh. The model is trained individually for each patient. This section provides a comprehensive description of the components of the approach, including the 3D organ representation, model architecture, and loss functions.

\begin{figure*}[ht]
\centerline{\includegraphics[width=450px]{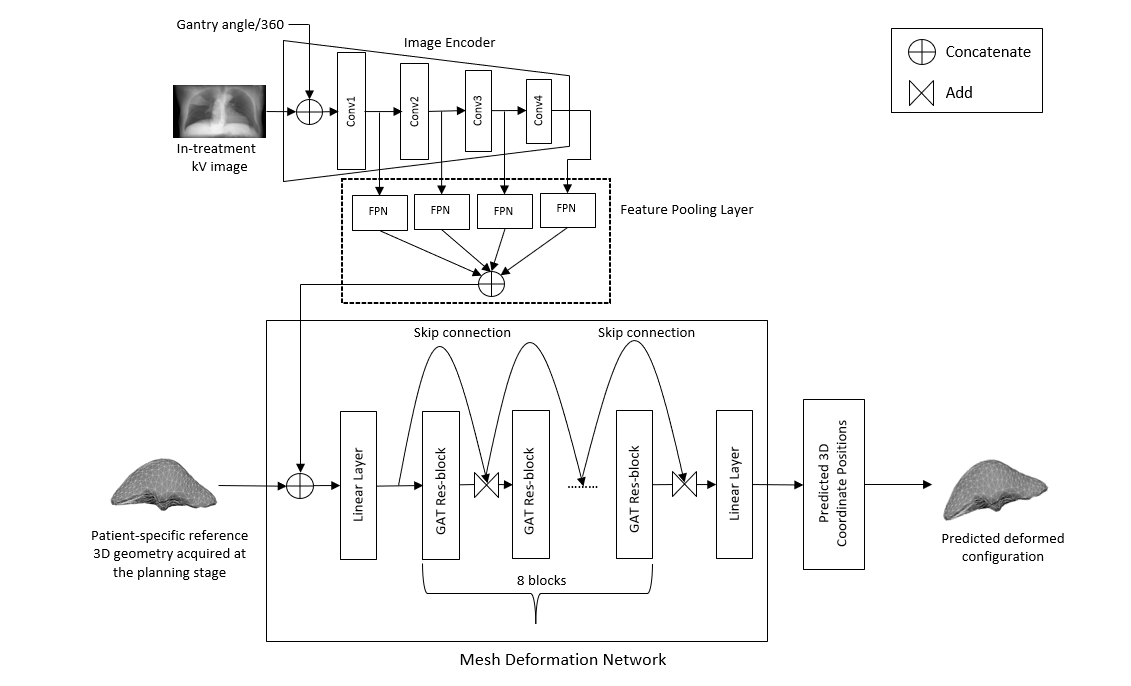}}
    \caption{Illustration of the Deep-Motion-Net architecture. A 2D-CNN image encoder extracts projection angle-dependent semantic features from an input kV X-ray image. A feature pooling layer comprising four learnable feature pooling networks attaches these features to the appropriate vertices in the patient-specific template mesh. Finally, a graph-attention-based network predicts the corresponding mesh deformation.}
    \label{fig:model_architecture}
\end{figure*}

\subsection{3D organ shape representation} \label{Sect:ShapeRep}

We use a 3D unstructured tetrahedral mesh to describe the volumetric organ shape, rather than its surface only. The mesh can be described as an undirected graph $G = \{V, E, F\}$, where $V$ is the set of $N$ vertices in the mesh, $E$ represents the set of edges between connected vertices, and $F$ are feature vectors attached to vertices. Patient-specific organ template meshes, derived from reference CT volumes, are constructed for each patient. In our experiments we used meshes with $N = 785$, $827$, $803$, and $756$, respectively, for livers in four patients.

\subsection{Model architecture}
The proposed architecture (shown in Figure \ref{fig:model_architecture}) consists of two components: a 2D-CNN encoder with four learnable feature pooling networks (FPNs), which extract perceptual features from the kV image and attach them to mesh nodes and a GAT-based mesh deformation network.

\subsubsection{Incorporating projection angle}
Inputs to the 2D-CNN encoder are a single-view kV image and its corresponding projection angle. We first normalise the projection angle by dividing it by 360 since the gantry rotation varies from 0 to 360 degrees. Before sending the input image to the image encoder, we concatenate the projection angle information as an extra channel. This concatenation part is accomplished by expanding and reshaping the projection angle to align with the spatial resolution of the input image (i.e. 256$\times$256). 

\subsubsection{2D-CNN configuration}
The projection angle-dependant perceptual features of the image are then extracted using four convolutional layers in the image encoder, containing 16, 32, 64, and 128 filters, respectively. For all convolutional layers, the kernel size was set to 3$\times$3, and stride and padding were both set to 1. Exponential Linear Units (ELUs) were applied for non-linearity after every convolutional operation. Hidden layer outputs were normalised using group normalisation (\cite{wu2018group}) since this helps reduce the internal covariate shift, which regularly alters the distribution of the hidden-layer activations during model training. Output feature maps of each convolutional layer were then down-sampled using 2$\times$2 max-pooling layers with stride two before passing into the next layer.

\subsubsection{Feature pooling networks}
We integrated learnable feature pooling networks (FPNs) to efficiently learn optimal associations between image features and mesh nodes in our architecture. This utilization of FPNs allows the network to maximise the predictive performance through end-to-end training, eliminating non-trainable components and enabling full optimization of feature learning capabilities for the downstream graph network. We use four FPNs, each coupled with its respective CNN convolutional layer (Figure \ref{fig:model_architecture}). Each FPN has two layers: an adaptive average pooling layer (AAP) and a FC layer. The AAP, with output size 7$\times$7, is applied over the output of the corresponding convolutional layer to reduce the dimension so that an input feature map with dimensions $H\times W\times C$ is reduced to $7\times7\times C$. The output is flattened before sending it into the FC layer, which contains $5N$ neurons. We reshaped the output feature vector of each FPN into ($N$, 5). These outputs were then concatenated with 3D coordinates of the template mesh before feeding into the graph network. Each mesh vertex thereby acquires a total of 23 features: five from each of the four FPNs, and three representing the 3D vertex position.

\subsubsection{Mesh deformation network}
The objective of this network is to estimate the 3D coordinates for each vertex in the deformed mesh configuration. We developed a GNN architecture which relies on a series of GAT-based blocks with residual connections (i.e. skip/shortcut connections). We applied residual connections to increase the impact of earlier layers on the final node embeddings. These connections substantially speed up training and produce better quality output shapes. The fundamental building block is identical to the Bottleneck residual block (\cite{he2016deep}), with 1$\times$1 convolution and 3$\times$3 convolution layers replaced by per vertex FC layers and GAT layers (\cite{velivckovic2017graph}), respectively. Further, group normalization layers are used instead of batch normalization (\cite{ioffe2015batch}) since small batch sizes result in incorrect estimates of batch statistics, which substantially increases model error. We employed ELUs to impart non-linearity.

This deformation network is similar to the graph-CNN architecture described in \cite{kolotouros2019convolutional}, but with two differences: graph convolutional network (GCN) layers are replaced by GAT layers and ReLU is replaced by ELU activation. GAT layers incorporate attention mechanisms which allow the assignment of different weights for different neighboring node feature vectors depending on how they interact. GCNs (\cite{kipf2016semi}), by contrast, use isotropic filtering and thereby assign similar weight to all feature vectors around the current node. ELU activation avoids the dying ReLU problem and may generate negative non-linear outputs.

\subsection{Loss functions}

The overall objective function $\mathcal{L}$ is defined as:

\begin{equation} 
\label{eq:total_loss}
\mathcal{L} = \mathcal{L}_{Shape} + \lambda \mathcal{L}_{Laplacian},
\end{equation}
with weighting term $\lambda$. $\mathcal{L}_{Shape}$ quantifies the difference between the predicted and ground-truth 3D meshes. The template mesh starting shape is defined by its vertices $V$ (Sect.~\ref{Sect:ShapeRep}). We define $Y$ and $\hat{Y}$ to be the ground-truth and predicted deformed positions of these vertices. An intuitive objective is then to minimize the per vertex $L_{1}$ loss between $Y$ and $\hat{Y}$:

\begin{equation} 
\label{eq:l1_loss}
\mathcal{L}_{Shape} = \sum_{i=1}^{N} \norm{\hat{Y}_{i} - Y_{i}}_{1},
\end{equation}
where $Y_{i}$ and $\hat{Y}_{i}$ are the $i^{\rm{th}}$ vertices in the respective sets.

Using $\mathcal{L}_{Shape}$ alone the vertices were found to move too freely. We introduced a discrete Laplacian (\cite{wang2018pixel2mesh}) loss $\mathcal{L}_{Laplacian}$ as a regularization term to limit this freedom by ensuring vertices do not move too far in relation to their neighbours. The additional constraint ensures that the mesh has a smooth surface. The discrete Laplacian of a vertex $\hat{Y}_{i}$ is denoted as:

\begin{equation}
    \delta_{\hat{Y}_{i}} = \frac{1}{|S(\hat{Y}_{i})|} \sum_{j\in S(\hat{Y}_{i})} (\hat{Y}_{i} - \hat{Y}_{j}),
\end{equation}
where  $\hat{Y}_{j}$ is a neighbouring vertex of $\hat{Y}_{i}$ and $S(\hat{Y}_{i})$ denotes the set of all such neighbours. The discrete Laplacian loss is then given by:

\begin{equation}
    \mathcal{L}_{Laplacian} = \frac{1}{N} \sum_{i=1}^{N} \norm{ \delta_{V_{i}} - \delta_{\hat{Y}_{i}} }_{2}^{2},
\end{equation}
where $\delta_{V_{i}}$ and $\delta_{\hat{Y}_{i}}$ are the discrete Laplacian before and after the deformation, respectively.

In our experiments, we used $\lambda=0.1$ to balance the weights of the two losses.

\subsection{Implementation and training details}

The complete network was implemented using PyTorch and PyTorch-geometric (\cite{Fey_Lenssen_2019}). The model was trained using the Adam optimizer, with a learning rate of 0.0002 and weight decay of 0.001. The batch size for both training and validation datasets was set to 16. Group normalization was used to normalize the hidden layer outputs of both 2D-CNN and GNN. We used single-headed attention in the GAT layers due to memory restrictions. Early stopping was employed to monitor the validation loss with a patience of 30 consecutive epochs, and if the loss did not improve, the optimizer itself stopped the training. However, if there was no progress after eight consecutive epochs, the learning rate was decreased by a factor of 0.8. All weights were initialized using the scheme in \cite{glorot2010understanding}. The gradient descent converged after 400 epochs to the optimal solution. As described in \ref{optimisation_study}, we conducted a series of experiments to determine the optimal network components. The training lasted approximately 36 hours with 256$\times$256 image resolution on a Nvidia Quadro RTX 4000 GPU using a Precision 7820 Tower XCTO Base workstation.

\section{Synthetic dataset generation}\label{dataset}

Model training requires paired sets of organ motion instances and corresponding kV images. However, to the best of our knowledge, there are no means of directly measuring such motions while also acquiring the requisite images. Hence, we use synthetically generated data to train and, in this work, to evaluate the model. Plausible patient-specific motion patterns are extracted from 4D-CT images, and new synthetic instances are produced by interpolating and, within reasonable bounds, extrapolating from these. The process is as follows: 1) 4D-CT images are analysed using the SuPReMo toolkit (Surrogate Parameterized Respiratory Motion Model -  \cite{mcclelland2017generalized,jamieSupremo2017}), which produces, among other things, a model of the motion present in the images, linked with appropriate surrogate signals; 2) new, yet plausible motion instances are generated from this model by randomly perturbing the surrogate signal; 3) the resulting motion fields are used to deform the reference (phase 0) CT volume; 4) Digitally Reconstructed Radiographs (DRRs) are generated from these deformed volumes for all required projection angles, and 5) the DRRs are style transferred to match kV image intensity and noise distributions. The result is a set of realistic `kV' images of the deformed anatomy acquired at various projection angles, for which ground-truth 3D motion states are known. Finally, the target organ is segmented from the reference CT volume, and an organ template mesh is constructed. Full details are presented in the following sections.

\subsection{Generation of synthetic motion states from 4D-CT data}
\label{sectSynthMotion}

\begin{figure*}[ht]
\centering
\subfloat{\includegraphics[width=160px]{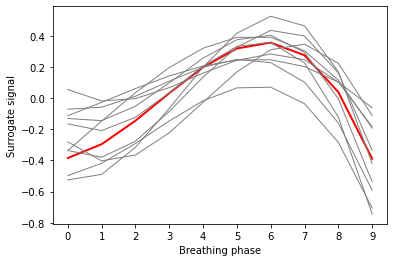}}
\qquad
\subfloat{\includegraphics[width=160px]{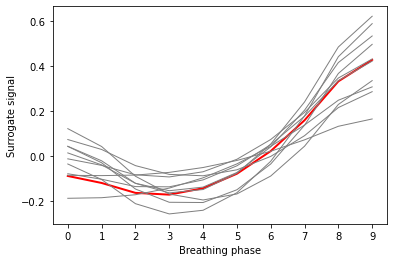}}
\qquad
\caption{Example surrogate signals: original signals associated with the input 4D-CT data (red) and randomly generated variations from these (grey), used in turn to synthesise new motion states. The first and second signals are plotted at the left and right, respectively.}
\label{fig:surrogate_signals_and_variants}
\end{figure*}

SuPReMo is a toolkit for simultaneously estimating a motion model and constructing motion-compensated images from a 4D-CT dataset. The resulting model describes the anatomical motion present in the raw data over a single (averaged) breath cycle and is linked with corresponding scalar surrogate signals derived, for example, from breathing traces or image features. For formulation reasons (see \cite{mcclelland2017generalized,jamieSupremo2017}), two surrogate signals are required; however, these needn't be independent, and as a practical measure in each case we constructed the second signals as temporal derivatives of the first, computed for example using finite differences. Example surrogate signals are plotted in Fig.~\ref{fig:surrogate_signals_and_variants}. Our 4D-CT datasets each decompose the breath cycle into 10 bins (phases).

First, we fit SuPreMo's motion model for each individual case with input surrogate signal and 4D-CT data. This returns the motion-compensated reconstructed 3D-CT (MCR) image volume and the fitted motion model, which are then utilised to simulate new motion states by varying the input surrogate signal. Each point $s_i, \, i \in [0, 9]$ on the curve is randomly perturbed by a value in the range $\pm 0.4 s_i$. Extrapolating beyond this range is more likely to cause unrealistic motion and even folding in the resulting images. A new surrogate signal is then created by fitting a $3^{\rm rd}$ order polynomial to the new points. The latter ensures the new signal remains smooth over the full breath cycle. Examples of such generated signals are shown in Fig.~\ref{fig:surrogate_signals_and_variants}. Using this new signal, and MCR as a reference volume, SuPReMo's motion model generates corresponding deformed 3D-CT volumes and their related deformation vector fields. 

For each test case (Sect.~\ref{evaluation_results}), we created 11 separate surrogate signals, each comprising ten deformed configurations, resulting in 110 synthetic deformation states in total. To introduce more diversity into the synthetic motion instances and substantially deviate from the original 4D-CT data, we incorporated rigid motions by applying random translations/shifts along the left-right (LR), anterior-posterior (AP), and superior-inferior (SI) directions and produced a total of 550 deformed states. This is also advantageous during testing on real in-treatment kV images, as it reflects the variations in onboard patient setup across different scan series or fractions, potentially leading to shifts in the in-treatment kV images.

\subsection{Generation of synthetic kV X-ray images}

For each deformation state, DRRs were obtained from the deformed 3D-CT volumes using the Siddon-Jacobs ray tracing technique (\cite{jacobs1998fast,siddon1985fast}). The latter, given a source position and projection direction, computes the line integral of Hounsfield unit densities along ray lines to a prescribed 2D plane. In radiotherapy, in-treatment kV images are acquired perpendicularly to the anatomical axial direction.  

The resulting DRRs lack scatter properties and noise characteristics of genuine kV X-ray images, making them appear sharper and higher in contrast. We, therefore, used a CycleGAN (\cite{cycle_gan_zhu}) by conditioning on projection angle to effect DRR-to-kV style transformation. The conditional CycleGAN was trained on an unpaired set of DRRs and real in-treatment kV images for each case separately since the FOV acquisition varies from patient to patient. Histogram equalization was first applied on the kV images to stretch the intensity histograms concentrated on a narrow interval to a broader range and increase contrast. All DRRs were then passed to the conditional CycleGAN to create the final synthetic kV X-ray images.

\subsection{Creation of template meshes}

Binary masks of the relevant organs were extracted from reference 3D-CT volumes using 3D Slicer's Segment Editor. These masks were then used to generate tetrahedral meshes using the Iso2Mesh(~\cite{fang2009tetrahedral}) tool in MATLAB. The resulting meshes were scaled and translated to reflect image voxel sizes and correct image origins. By these means, the meshes are physically aligned with the relevant anatomical regions in the 3D-CT volumes. Ground-truth positions of the mesh nodes for each of the generated deformation states are produced by interpolating the displacement vector field at the node positions.

\section{Model evaluation and results}
\label{evaluation_results}

\subsection{Clinical datasets}
We evaluated the Deep-Motion-Net framework using data from four liver cancer patients, with focus on liver motion. Each patient dataset comprised: 1) a 4D-CT with 10 phases (i.e. 10 $\times$ 3D-CT volumes), spatial resolutions of $0.98\times0.98\times2.0$ mm$^3$, and image dimensions of $512\times512\times105$; and 2) two kV scan series, each covering approx. 4-5 mins of free breathing and a full rotation of the treatment gantry, acquired at the start of treatment sessions on different days. As per clinical practice, the kV scans were centered on the liver region and used a constrained FOV. As a result, for some projection angles, the images include only a segment of the liver.

\subsection{Experiments on synthetic data}
\label{Experiments_synthetic}

We first evaluated the framework's ability to recover motion states synthetically generated from the clinical data, and for which ground-truth deformations were correspondingly available. For each patient, synthetic motion and image data were generated as per Sect.~\ref{dataset}. Each patient's 550 deformation states were then split into training, validation, and test sets in the proportions 350, 100, and 100, respectively. For training and validation deformation states, 100 uniformly sampled (i.e. at projection angle intervals of $3.6^\circ$) synthetic kV images were generated (giving 35,000 training and 10,000 validation images). For test states, 50 kV images were generated at \emph{randomly} sampled projection angles (giving 5,000 test images). To ensure good FOV alignment of the synthetic and real kV images for each patient, each synthetic image was generated using FOV origin header information from a corresponding (i.e. acquired at a comparable projection angle) real image within one of the scan series. To avoid doubt, the real kV images subsequently played no further role in the experiment. Model-predicted and ground-truth 3D liver shapes were compared for all synthetic kV images in the test sets.

\begin{figure*}[!ht]
\centering
\subfloat{\includegraphics[width=\linewidth]{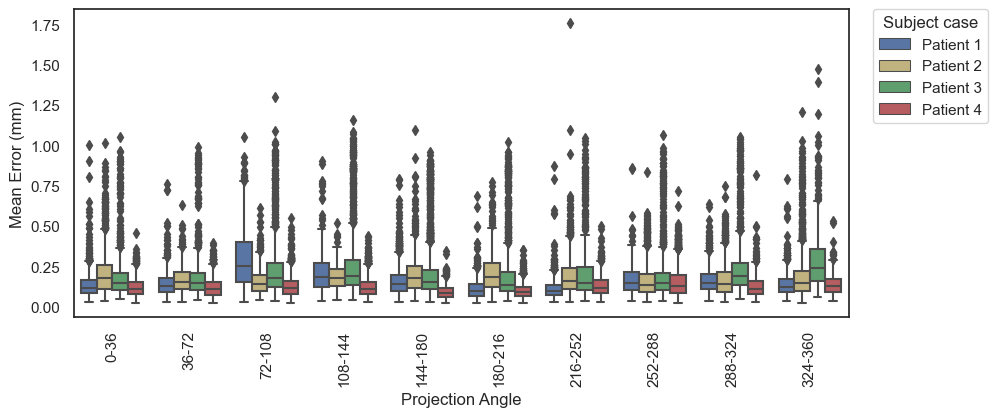}}

\subfloat{\includegraphics[width=\linewidth]{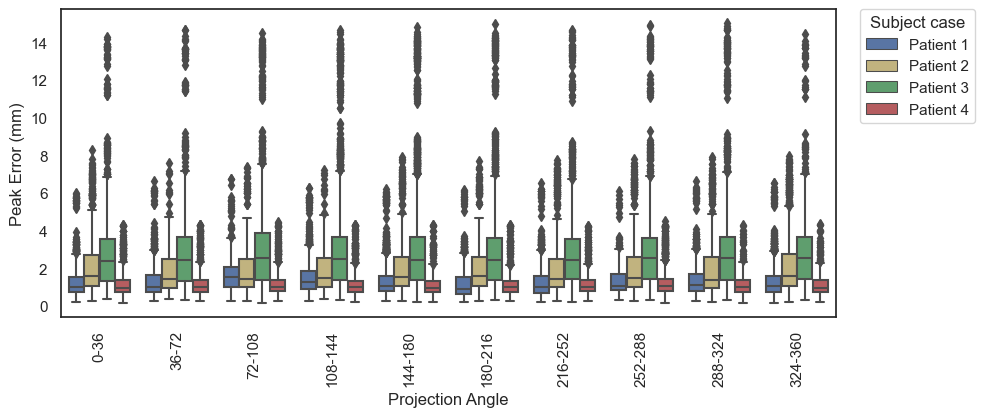}}

\caption{Effect of projection angle on prediction accuracy: box and whisker plots of mean (top) and peak (bottom) prediction errors grouped according to image projection angle (degrees). Each box and whisker shows the distribution of errors for the indicated projection angle using all deformation states in the test set. For clarity of visualisation, angles are further grouped into 10 equal bins covering a full revolution. Results for patients 1 (blue), 2 (yellow), 3 (green), and 4 (red) are shown for each bin.}
\label{fig:mean_peak_errors_projection_angle_all_subjects}
\end{figure*}

\begin{figure*}[!ht]
\centerline{\includegraphics[width=\linewidth]{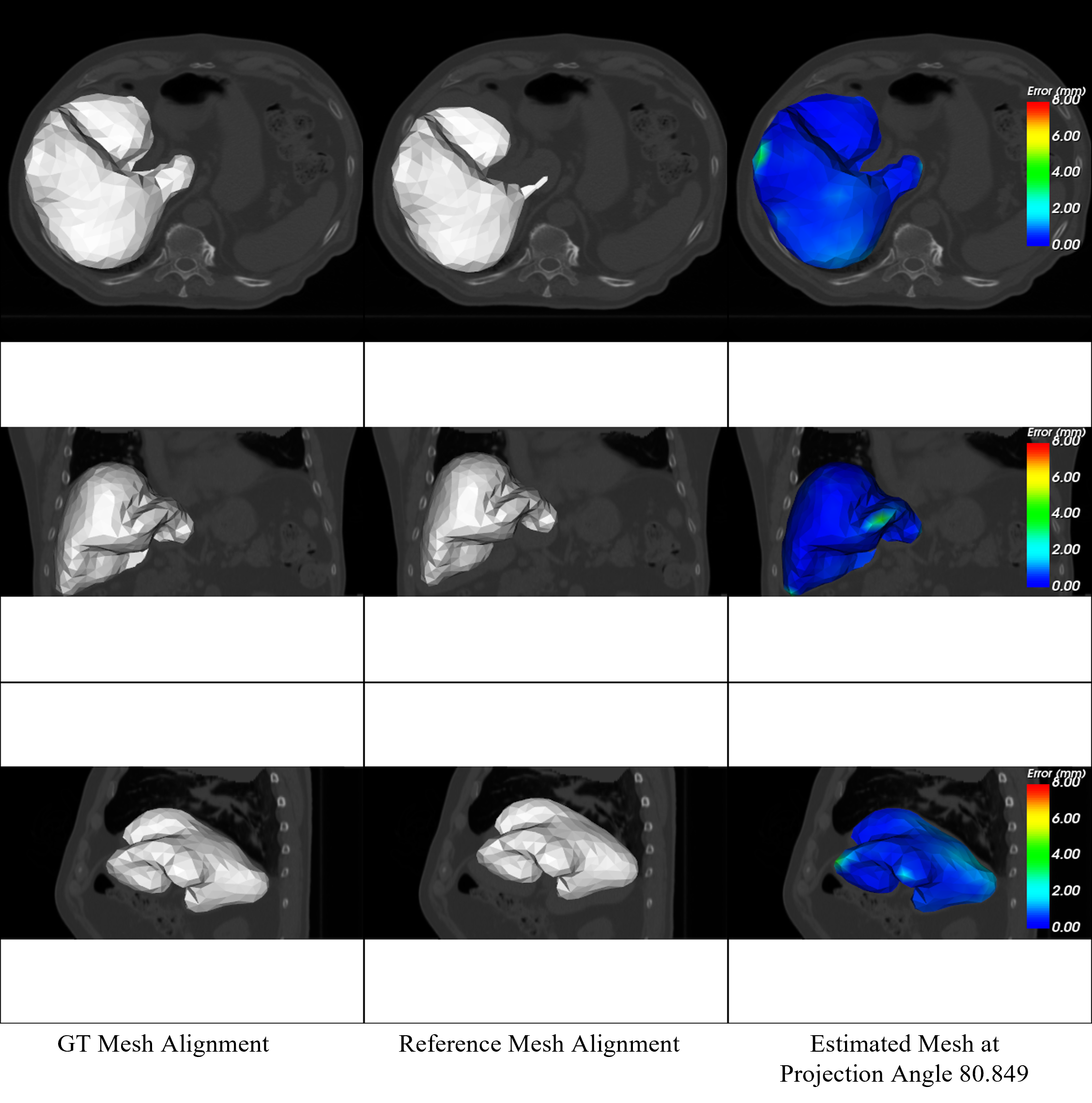}}
    \caption{Visualisations of ground-truth deformed (left column), template (middle column), and estimated deformed (right column) 3D liver shapes. Meshes are overlaid on the deformed 3D-CT volume. Rows 1-3 show, respectively, axial, coronal, and sagittal views. Results are shown for the \emph{worst} performing test case for patient 1: image projection angle $80.849^\circ$, and deformation state producing highest errors. Contours in the right column indicate the spatial distribution of errors on the surface. Similar results for patients 2-4 are presented in the supplementary material.}
    \label{fig:liver_mesh_alignment_visualization_patient1}
\end{figure*}

Summary results are presented in Table \ref{table:summary_stat_table}. The study employed Euclidean distance as the metric to evaluate the distance errors between ground-truth and estimated shapes. Distributions of mean and peak errors for each test set, and for each projection angle (divided into bins), are shown in Figure \ref{fig:mean_peak_errors_projection_angle_all_subjects}. Finally, surface renderings of the ground-truth, reference, and predicted mesh shapes, overlaid on deformed 3D-CT volumes, are presented in Figure \ref{fig:liver_mesh_alignment_visualization_patient1}.

\begin{table*}[ht]
\begin{center}
\caption{Summary statistics for all test sets with synthetic kV images. Patient case numbers are indicated in column 1. $E_{pred}$ and $U_{GT}$ refer to prediction errors and underlying ground-truth nodal deformation magnitudes, respectively. \emph{Mean (std)}: means (and standard deviations) of values across all nodes, all deformation states, and all projection angles. \emph{Mean peak}: means of the peak values for each deformation state across all projection angles. \emph{Max peak}: overall maximum values from all nodes, deformation states and angles. \emph{99$^{th}$ Percentile}: $99^{\rm th}$ percentile values from all nodes, deformation states and angles. All values reported in mm.}
\label{table:summary_stat_table}
\begin{tabular}{|c|c|c|c|c|c|}
\hline
\textbf{Case} & \textbf{} & \textbf{Mean (std)} & \textbf{Mean peak} & \textbf{Max peak} & \textbf{$99^{\rm th}$ Percentile} \\
\hline
\multirow{2}{*}{1} & $E_{Pred}$ & 0.16$\pm$0.13 & 1.39 & 6.75 & 0.75 \\ \cline{2-6} 
 & $U_{GT}$ & 10.18$\pm$1.33 & 14.71 & 28.12 & 13.85\\ \cline{1-6} 
\multirow{2}{*}{2} & $E_{Pred}$ & 0.18$\pm$0.19 & 1.99 & 7.97 & 0.97\\ \cline{2-6} 
  & $U_{GT}$ & 11.65$\pm$1.76 & 15.76 & 34.91 & 14.38\\ \cline{1-6} 
\multirow{2}{*}{3} & $E_{Pred}$ & 0.22$\pm$0.34 & 3.29 & 14.66 & 1.81\\ \cline{2-6} 
  & $U_{GT}$ & 14.89$\pm$2.54 & 19.36 & 49.63 & 17.65\\ \cline{1-6} 
\multirow{2}{*}{4} & $E_{Pred}$ & 0.12$\pm$0.11 & 1.16 & 4.36 & 0.51\\ \cline{2-6} 
  & $U_{GT}$ & 10.07$\pm$1.13 & 12.86 & 25.64 & 11.97\\ \hline
\end{tabular}
\end{center}
\end{table*}



For each of the four test sets the overall mean error was low: $\leq0.22$ mm. Within each test set, the maximum peak errors (i.e. overall max error found in any of the deformation states and at any node) were rather higher, ranging from 4.36 mm for patient 4 to 14.66 mm for patient 3. These peak values occurred for the deformation states with the highest ground-truth displacement (namely: 28.12, 34.91, 49.63, and 25.64 mm for patients 1, 2, 3, and 4, respectively) for each test set. Higher underlying ground-truth displacements corresponded, \emph{in general}, with higher peak errors, although mean errors were consistently low. As indicated by Figure \ref{fig:mean_peak_errors_projection_angle_all_subjects}, the displacement prediction accuracy was almost independent of the image projection angle: box-and-whisker plots of both mean and peak errors are very similar across the range of angles.

Finally, it is important to note that the higher peak errors in all cases were extremely localised within the meshes. This is indicated firstly by the low overall mean values and more so by the $99^{\rm th}$ percentile errors (Table \ref{table:summary_stat_table}), which were below 1 mm for 3/4 test sets and below 2 mm for the fourth. That is, the errors were low, even sub-millimetres, for the vast majority of the mesh nodes. The point is further illustrated by the renderings of predicted mesh shapes colour-mapped by displacement error in the right columns of Figure \ref{fig:liver_mesh_alignment_visualization_patient1}. For liver meshes, the error was close to zero over most of the surface, and had only very localised regions of higher values.

\subsection{Evaluation on real kV images}
\label{evaluate_realkV_images}
Our second set of experiments used real in-treatment kV images from each patients' second scan series (i.e. the series \emph{not} used during training data creation). These series contained 1378, 1310, 1320, and 1292 images for patients 1, 2, 3, and 4, respectively. In the absence of ground-truth deformations, direct assessment of prediction errors is impossible. Therefore, two approaches were adopted: 1) semi-quantitative assessment based on an image similarity metric between input real kV images and model-generated DRRs; and 2) qualitative assessment based on overlaying model-predicted liver boundaries on input kV images. For the qualitative assessment, all images in the scan series were used. To reduce computation time (associated, in particular, with spline deformation of the image volumes), only 100 images, uniformly sampled, were used from each patients' series in the similarity-based assessment.

\subsubsection{Mutual Information-based assessment:}

3D organ deformations predicted for a given input kV image can be used to deform the patient's reference CT volume. The correspondence between the input image and a DRR generated from this deformed CT volume should then improve with the accuracy of the model prediction. With this in mind, we used kV-to-DRR image similarity, quantified using mutual information (MI), as a surrogate measure of the model's deformation prediction accuracy. In particular, we assessed the improvement in MI when using the model-deformed CT volume compared with using the undeformed volume.

\begin{figure*}[ht]
\centerline{\includegraphics[width=\linewidth]{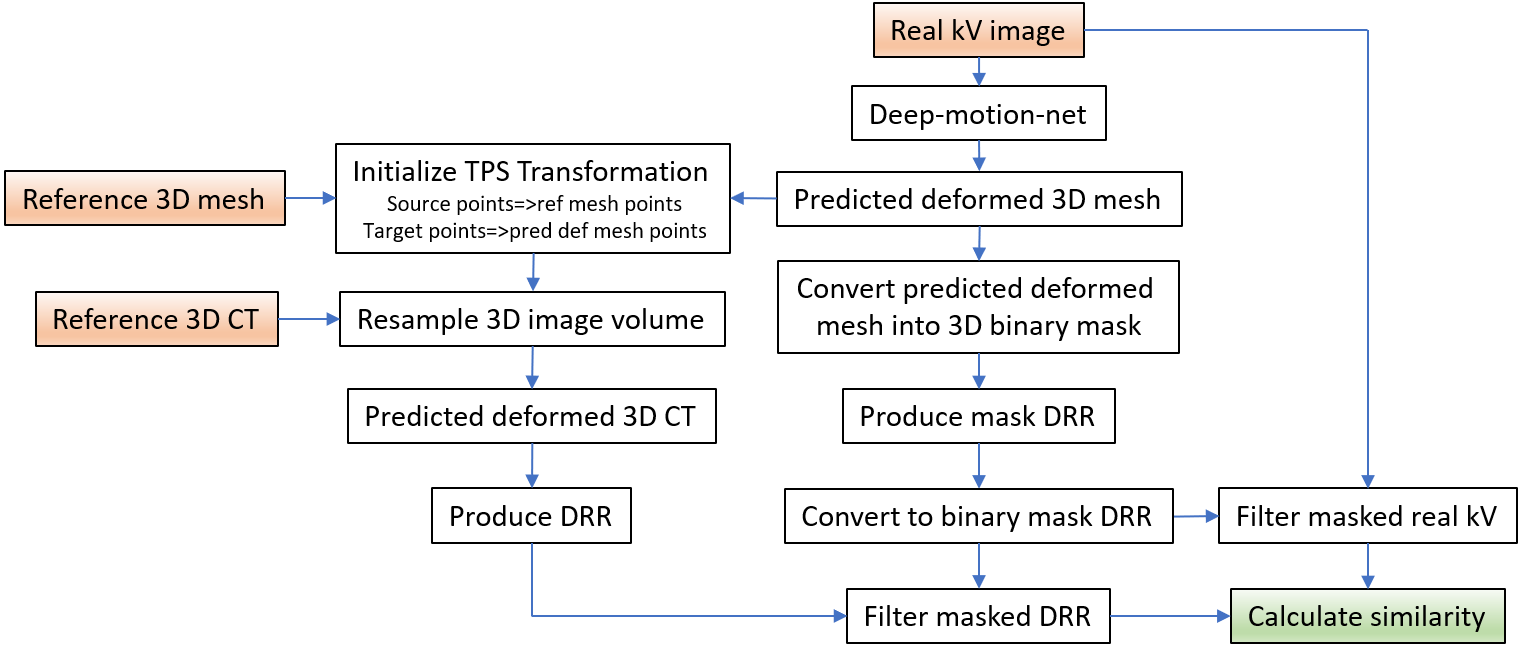}}
    \caption{Illustration of the process of MI-based assessment of model prediction accuracy.}
    \label{fig:real_kv_evaluation}
\end{figure*}

The process is summarised in Figure \ref{fig:real_kv_evaluation}. The input kV image is passed to the model, which predicts the corresponding 3D organ mesh deformation. A thin-plate-spline (TPS) transformation is initialised using the reference and deformed mesh, and used to deform the reference 3D-CT volume. A DRR is then generated using the input kV image's projection angle. Separately, the deformed mesh is used to generate a 3D binary mask, covering the liver-predicted shape, from which a 2D `mask DRR' is produced. The latter is used to create regions of interest (ROIs) in both the DRR described above and the input kV image. MI is then computed between these two ROIs. The ROI masking process ensures the similarity computation is restricted to the image regions to which the model predictions apply; deformations in the remainder of the 3D-CT volume, derived from the TPS transformation, are merely extrapolated rather than directly predicted. Finally, a comparable process is followed to produce a reference DRR (i.e. without accounting for the motion) for comparison: the DRR is generated from the undeformed (reference) 3D-CT volume, a ROI mask is created from the reference organ mesh, and MI is computed between the masked DRR and input kV image.


Table \ref{table:mutual_info_stat_table} summarizes the results of these experiments. The similarity score values using the reference 3D-CTs are lower than for the motion-corrected (deformed) volumes, suggesting our model is making sensible predictions of the liver motion. Moreover, the marginal disparities in average MI values led us to conduct a one-way ANOVA test for each patient case. For this purpose, we generated two groups. In the first group, MI scores were computed by comparing masked DRRs generated using predicted deformed 3D-CT image volumes with corresponding masked real kV images. The mask was produced using the mask DRR projection derived from the predicted deformed 3D binary mask volume. In the second group, MI scores were recalculated, this time by comparing the DRR generated from the undeformed (reference) 3D-CT volume with the input kV image. Here, the ROI mask was created from the reference organ mesh. Subsequently, we conducted a one-way ANOVA test based on these two groups for each test dataset. The resulting p-values for MI differences between the reference and deformed versions were 0.0449, 0.0283, 0.0453, and 0.0423 for patients 1, 2, 3, and 4, respectively, suggesting statistically significant (assuming alpha value of 0.05) differences.

\begin{table}[ht]
\begin{center}
\caption{MI similarity scores (mean $\pm$ standard deviation, computed from the 100 images sampled from each scan series) between real kV images and DRRs generated at the same projection angles for each patient case. Column 2 presents values when the reference (i.e. undeformed) CT volume is used. Column 3 presents values when the CT volume is deformed using the model-predicted deformation fields. All values were computed on the liver region.}
\label{table:mutual_info_stat_table}
\begin{tabular}{|c|c|c|}
\hline
\textbf{Case} & \textbf{Reference} & \textbf{Deformed}\\
\hline
1 & 1.16$\pm$0.31 & 1.33$\pm$0.23 \\ \cline{1-3} 
2 & 1.14$\pm$0.21 & 1.39$\pm$0.14 \\ \cline{1-3} 
3 & 1.13$\pm$0.27 & 1.28$\pm$0.23 \\ \cline{1-3} 
4 & 1.31$\pm$0.25 & 1.47$\pm$0.15 \\ \hline
\end{tabular}
\end{center}
\end{table}

\subsubsection{Qualitative assessment by boundary overlay:}

\begin{figure*}[!ht]
\centering
\subfloat{\includegraphics[width=\linewidth]{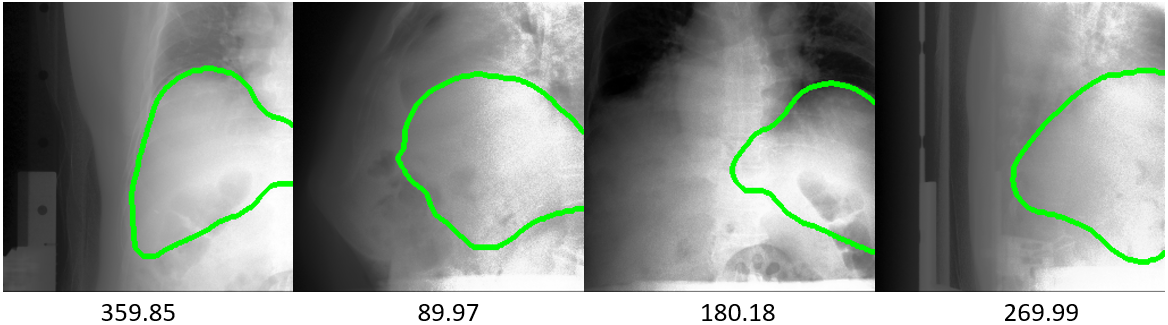}}\\[-0.2ex]

\subfloat{\includegraphics[width=\linewidth]{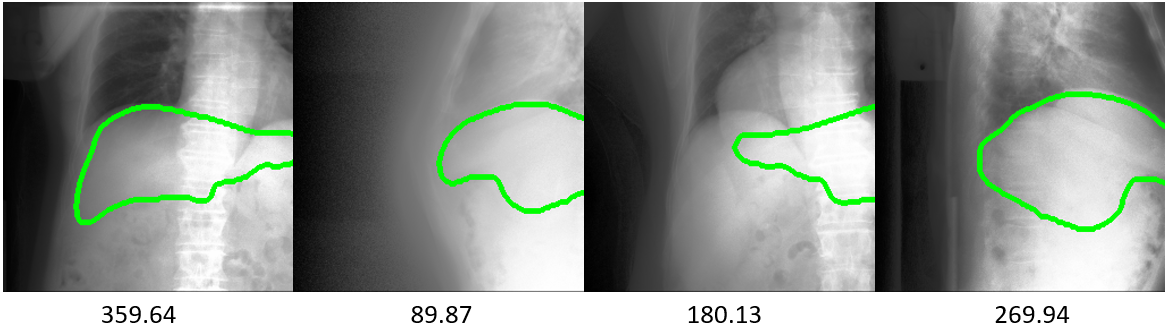}}\\[-0.2ex]

\subfloat{\includegraphics[width=\linewidth]{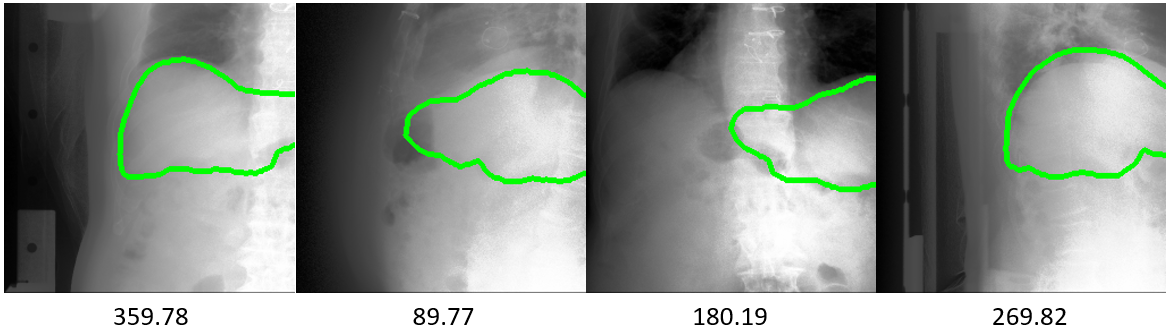}}\\[-0.2ex]

\subfloat{\includegraphics[width=\linewidth]{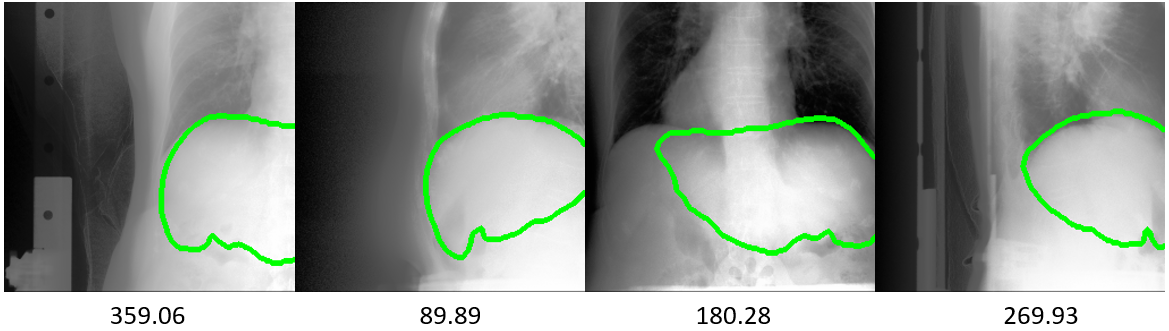}}

\caption{Samples of overlaid predicted liver boundary projections on corresponding real kV images for the four patients. Rows 1-4 show, respectively, results for patients 1-4. Results for images acquired at four projection angles (degrees, indicated below the images) are shown.}
\label{fig:liver_motion_overlay_realkv}
\end{figure*}

Histogram-equalised kV images from the mentioned scan series were fed into the trained models to obtain 3D mesh shape predictions. As mentioned, all images in the scan series were used. From each predicted mesh, a corresponding binary image volume was generated. Finally, the projected liver surface boundaries were obtained through ray-tracing on these binary image volumes, and superimposed on the respective input kV images. Samples from each patient are shown in Figure \ref{fig:liver_motion_overlay_realkv}. Supplementary materials, moreover, include animated versions based on the full scan series of each patient case for further reference. These latter most effectively show (qualitatively) the consistent alignment of the model predictions with the input images across many breathing cycles and the full rotation of the treatment gantry.






\subsection{Comparison model}

We compared the performance of our approach with that of the recently presented IGCN model (\cite{nakao2021image,Nakao_9844010}). As described, like our approach, IGCN predicts 3D organ shapes from single-view X-ray images. It is, however, limited to images with a constant projection angle (e.g. 0$^\circ$, corresponding to anterior-posterior projection), and predicts only 3D surface geometries rather than 3D volumetric configuration. To ensure fairness, we therefore conducted the comparison on this basis.

We trained our model as previously described, and using the data described in sect. \ref{Experiments_synthetic}. To train the IGCN model, we first extracted surface meshes from the ground-truth volumetric meshes in the training, validation, and test sets. We then trained the model using the deformed surface meshes and associated synthetic kV images for projection angle of zero. Image dimensions were 256$\times$256. Initial batch size, learning rate, and total number of epochs were as specified in \cite{nakao2021image}: 1, 0.0001 and 1000, respectively.

Images with projection angle zero from the test set were passed to the IGCN model, which predicted corresponding deformed surface meshes. To facilitate a meaningful comparison with our model, therefore, we derived corresponding surface meshes from the latter's predicted volumetric meshes for the same images. The outcomes presented in Table \ref{table:summary_comparision_table} for the test set show that our method achieved overall higher accuracy in liver surface deformation across all patient cases. Further, we conducted one-way ANOVA tests for each test set to check for statistically significant differences in mean errors between our approach and IGCN. Two groups were produced by calculating the mean of the Euclidean distances between ground-truth and predicted surface meshes independently for each test dataset, one for our model and the other for the IGCN model. Each list comprises 100 mean error values, reflecting the 100 deformed states in the test set, where each deformed state corresponds to a single kV (front-view) projection. Subsequently, we performed a one-way ANOVA test based on these two groups for each test dataset. The resulting p-values of 0.0419, 0.0088, 0.0074, and 0.0485 for patients 1, 2, 3, and 4, respectively, confirmed this significance (assuming 0.05 alpha value).

\begin{table*}[ht]
\begin{center}
\caption{Summary prediction error statistics from performance comparison between our model and IGCN. All errors are computed with respect to predicted organ surface meshes. Patient case numbers are indicated in column 1. $E_{Pred}^{Ours}$ and $E_{Pred}^{IGCN}$ refer to prediction errors for our method and IGCN, respectively. $U_{GT}$ refer to underlying ground-truth deformation magnitudes. \emph{Mean (std)}: means (and standard deviations) of values across all nodes, all deformation states, and all projection angles. \emph{Mean peak}: means of the peak values for each deformation state across all projection angles. \emph{Max peak}: overall maximum values from all nodes, deformation states and angles. \emph{99$^{th}$ Percentile}: $99^{\rm th}$ percentile values from all nodes, deformation states and angles. All values reported in mm.}
\label{table:summary_comparision_table}
\begin{tabular}{|c|c|c|c|c|c|}
\hline
\textbf{Case} & \textbf{} & \textbf{Mean (std)} & \textbf{Mean peak} & \textbf{Max peak} & \textbf{$99^{\rm th}$ Percentile} \\
\hline
\multirow{3}{*}{1} & $E_{Pred}^{Ours}$ & 0.17$\pm$0.11 & 0.89 & 4.91 & 0.47 \\ \cline{2-6} 
 & $E_{Pred}^{IGCN}$ & 0.18$\pm$0.25 & 1.13 & 6.37 & 0.93\\ \cline{2-6} 
 & $U_{GT}$ & 10.11$\pm$1.24 & 13.94 & 28.12 & 13.53\\ \cline{1-6} 
\multirow{3}{*}{2} & $E_{Pred}^{Ours}$ & 0.14$\pm$0.15 & 1.53 & 7.13 & 0.67\\ \cline{2-6} 
 & $E_{Pred}^{IGCN}$ & 0.17$\pm$0.21 & 2.81 & 8.79 & 1.18\\ \cline{2-6} 
 & $U_{GT}$ & 11.37$\pm$1.55 & 15.41 & 34.91 & 14.09\\ \cline{1-6} 
\multirow{3}{*}{3} & $E_{Pred}^{Ours}$ & 0.15$\pm$0.13 & 1.46 & 13.83 & 0.77\\ \cline{2-6} 
 & $E_{Pred}^{IGCN}$ & 0.19$\pm$0.23 & 2.05 & 14.31 & 1.23\\ \cline{2-6} 
 & $U_{GT}$ & 14.52$\pm$2.39 & 18.71 & 49.63 & 16.74\\ \cline{1-6} 
\multirow{3}{*}{4} & $E_{Pred}^{Ours}$ & 0.12$\pm$0.09 & 0.78 & 3.98 & 0.44\\ \cline{2-6} 
 & $E_{Pred}^{IGCN}$ & 0.14$\pm$0.17 & 0.97 & 5.31 & 0.77\\ \cline{2-6} 
 & $U_{GT}$ & 10.01$\pm$1.05 & 12.17 & 25.64 & 11.33\\ \hline
\end{tabular}
\end{center}
\end{table*}

\section{Discussion and conclusions}
We presented a GNN-based model for recovering 3D volumetric organ mesh deformation from a single in-treatment kV planar X-ray image acquired at arbitrary projection angles. This approach has several attractive features: it uses only readily accessible in-treatment imaging capabilities, rather than expensive and rare systems like MRI; it requires no extra sensing to provide surrogate signals; and no invasive fiducial marker implantation. Moreover, the model is end-to-end trainable, ensuring all components, and especially the image feature encoder, are optimised with respect to the overall prediction accuracy. To the best of our knowledge, this is the first example of a deep learning framework able to reconstruct volumetric 3D organ models accurately from arbitrary-angled single-view images, and thereby to enable such reconstructions across complete in-treatment scan series. We demonstrated the feasibility and accuracy of the technique using data from four liver cancer patients.

The model was trained with synthetic data constructed using the SuPReMo toolkit. Training in this way is essential in the absence of ground-truth deformations corresponding to real kV images; that is, there appears to be no other way of acquiring paired deformation/image sets for this scenario. For similar reasons, in this first study, direct quantitative evaluation of the model performance was also carried out using synthetic data. Naturally, the performance of the model will depend on the fidelity with which real patient motions are reproduced in the synthetic data. While SuPReMo appears to do a reasonable job of characterising the motion present in the input 4D-CT data, these data represent only averaged breathing cycles and do not by themselves provide information about the variability of this motion. Pragmatically, therefore, in the present work, we assumed some bounds on the variations from this average, and randomly generated motion states within these (Sect. \ref{sectSynthMotion}). We are exploring more rigorous approaches to characterising the patient motion variability (and incorporating this in model training data) based on kV image sequences acquired over several minutes during treatment.

While we focused here on respiratory motion estimation, our approach can in principle be used to predict any motion patterns, e.g. peristaltic motion of the gut or longer-term structural changes, given appropriate training data. As mentioned, the model involves no assumptions of periodicity or other specific motion characteristics. Clearly, the generation of training data is more challenging in some scenarios than in others, but this presents a practical difficulty rather than a theoretical one.

As mentioned, our approach has been developed specifically to accommodate input kV images acquired at arbitrary projection angles. We demonstrated this by training and evaluating the models with images generated at different projection angles. As shown in Figure \ref{fig:mean_peak_errors_projection_angle_all_subjects}, the prediction accuracy was indeed virtually independent of the projection angle.

We observed that the model prediction accuracy can be influenced by the quality and characteristics of the CT volumes from which training data are constructed. The higher peak errors found for patients 2 and 3 appear to be at least partly attributable to this. The CT volumes of these two patient cases exhibited certain reconstruction artefacts related to motion. For patient 3, in large amplitude deformed cases, these resulted in some mesh-image misalignment problems even in the ground-truth data (specifically, with the deformed states at initial and final time points) produced from SuPReMo. The deformed images for Patients 1 and 4, by contrast, contained no obvious artefacts or ambiguous anatomy, and peak errors were correspondingly smaller (Table \ref{table:summary_stat_table}). As emphasised, however, the regions of higher peak errors, even in patients 2 and 3, were nonetheless very localised, and mean errors remained low.

In the absence of ground-truth motion states for real in-treatment kV images, the evaluation was conducted through qualitative and semi-quantitative methods as described in Section \ref{evaluate_realkV_images}. Qualitative assessment involved visualizing overlaid organ contours on the input kV images, while semi-quantitative evaluation employed deforming reference CT volumes using predicted motions and calculating resulting image similarities. In the former experiment, the model appeared (again, qualitatively) to make sensible predictions. The animations included as supplementary material showed consistent and smoothly transitioning organ contours from one image to the next without noticeable jumps between frames, despite the prediction for each frame being independent of its neighbours (the model contains no recurrence mechanism, for example, and makes predictions based on single input images). This suggests the model is not greatly affected by noise or other irregularities present in the real images. In the latter experiment, compensating for the patient motion using the predicted deformation fields improved the similarity between the input images and corresponding model-derived DRRs, again indicating the model makes sensible predictions for the real images.

A limitation of our method, as currently implemented, is that we only estimate the deformation of the target organ; nearby OARs and other anatomy are ignored. However, the model itself imposes no restrictions in this respect. If suitable training data covering the relevant anatomy, and meshes of the relevant anatomical structures can be generated, the model can in principle estimate motions for these in the same way. SuPReMo, for example, provides deformation vector fields for the whole image volume, which could be used for this purpose. OAR and other meshes can normally be created similarly to the liver meshes used here. Potentially, the prediction of deformations for multiple disconnected meshes could result in anomalous overlapping regions. However, such instances would hopefully be rare when the training data include no such behaviour. If needed, separate non-overlap constraints in the formulation could also be conceived, which penalise mesh interpenetration in a way analogous to some contact formulations in computational mechanics. Albeit, this would increase the complexity of the approach. A simpler alternative could be deploying a single mesh covering all relevant anatomy. In all scenarios, it is possible that both model and training data size requirements would increase, though the overall approach would remain the same. 

A key part of the synthetic data creation was the development of a method for generating realistic synthetic kV X-ray images. While DRRs produced from ray tracing through CT volumes ultimately are also X-ray-based images, they do not suffer from the same scatter and noise phenomena of in-treatment kV X-ray images. Their appearance, consequently, is noticeably different. Therefore we first trained a CycleGAN \cite{cycle_gan_zhu} for each patient by conditioning on projection angle to learn the genuine kV X-ray style that can be transferred to the DRRs. CycleGANs are particularly well-suited to this task since they require only unpaired sets of DRRs and kV images. We initially used transposed convolution layers for the decoder part of the generators, however, we encountered checkerboard-like artefacts in several cases. These were eliminated by instead using convolution layers followed by pixel shuffle layers (\cite{Aitken_Ledig_Theis_Caballero_Wang_Shi_2017}).

The purpose of this research is to enhance the probability of curing cancer by maximising tumour control probability (TCP) while minimising normal tissue complication probability (NTCP). Conformal RT techniques, such as SABR, prove valuable in this context by enhancing TCP while keeping NTCP low, thereby increasing the Therapeutic Index. However, when there is motion, the healthy tissue located in the low-dose zone may migrate into the treatment field, leading to the delivery of a higher radiation dose to the healthy tissue. This, in turn, leads to patients enduring intolerable levels of toxicity. Another scenario arises when the tumour moves away from the radiation field, resulting in a reduced radiation dose to the tumour. Despite the consistent toxicity status, with the healthy tissue remaining stationary while the tumour moves, the tumour receives a suboptimal radiation dose, resulting in a lower TCP value. This necessitates a reduction in the prescribed dose to bring the NTCP back to an acceptable level. Hence, patients may not fully benefit from conformal radiation treatments like SABR due to challenges associated with motion. This, in turn, poses challenges for clinicians in delivering an optimal radiation dosage to patients, ultimately influencing the overall outcomes for patients. Therefore, with this research, we aim to ameliorate patient motion effects and improve the Therapeutic Index without the use of surrogate signals or invasive fiducial markers, enabling clinicians to reduce the NTCP to acceptable levels, thus allowing for larger prescription doses that will result in better outcomes for the patient. This will then enable clinicians to adapt therapies for diverse patient cohorts.

As mentioned, two general approaches to adapting RT treatments could be considered using motion estimates generated by our technique: inter- and intra-fraction adaption. Inter-fraction plan adaption could be achieved by retrospectively estimating the motion that occurred during a treatment fraction and computing motion-compensated spatial distributions of delivered dose (dose accumulation); the treatment plan could then be adapted accordingly for subsequent fractions. No new technology seems to be needed for this, beyond the requirement to image throughout the fraction, rather than during initial patient positioning only. We are currently developing a workflow for this process which, in outline, is as follows: 1) Deep-Motion-Net is trained using patient planning or treatment simulation 4D-CT data; 2) kV images are acquired throughout treatment delivery; 3) Deep-Motion-Net predicts organ motion states for all input kV images; 4) organ (mesh) motion predictions are used to deform patient reference 3D-CT volumes via thin plate spline transformation (with organ nodes as driving control points); 5) positions of the gantry/MLCs at the start and end of each time period are determined (interpolation in between); 6) dose for the predicted anatomy is computed; 7) dose is mapped to a common reference anatomy (corresponding to reference 3D-CT volume) 8) cumulative delivered dose (sum of contributions from individual motion states) is computed and compared with planned dose.

Intra-fraction adaption requires prediction of target motion, and adjustment of radiation delivery in response, in real-time during the treatment. While our model computes motions quickly (inference time for each input image is $\sim$27 msec), the inevitable response lag of the treatment system, due both to data processing following a motion input and the physical inertia of the delivery apparatus, means motion predictions must in practice be computed somewhat \emph{ahead of time}. In principle, our approach could be extended to enable this using ideas from sequence-to-sequence learning, e.g. by incorporating RNN (recurrent neural networks) or LSTM (long-short-term memory) components in the model to capture sequences of organ motions, given dynamic sequences of input kV images. Our recent work on generative modelling of cardiac MR image sequences (\cite{ZAKERI2023102678}) can provide further inspiration here.

\section*{CRediT authorship contribution statement}
\textbf{Isuru Wijesinghe:} Conceptualization, Methodology, Investigation, Software, Formal analysis, Writing – original draft. \textbf{Arezoo Zakari:} Software, Writing - review \& editing. \textbf{Alireza Hokmabadi:} Software. \textbf{Michael G. Nix:} Conceptualization, Resources, Writing – review \& editing, Supervision. \textbf{Bashar Al-Qaisieh:} Resources. \textbf{Ali Gooya:} Conceptualization, Writing – review \& editing, Supervision. \textbf{Zeike A. Taylor:} Conceptualization, Resources, Funding acquisition, Writing – review \& editing, Supervision.

\section*{Acknowledgements}
This work was supported by Cancer Research UK funding for the Leeds Radiotherapy Research Centre of Excellence (RadNet - C19942/A28832) and a Royal Academy of Engineering / Leverhulme Trust Research Fellowship (DIADEM-ART LTRF2021-17115). The authors gratefully acknowledge the support and advice of Dr Jamie McClelland on using SuPReMo.

\appendix

\section{Ablation study}
\label{optimisation_study}

\begin{table}[ht]
\begin{center}
\caption{Impact of reducing the number of image encoder MLPs. All values reported in mm.}
\label{table:impact_of_MLPs}
\begin{tabular}{|c|c|c|c|c|}
\hline
\textbf{Experiment} & \textbf{Mean (std)} & \textbf{Mean peak} & \textbf{Max peak} & \textbf{$99^{\rm th}$ Percentile}\\
\hline
1 & 0.21$\pm$0.18 & 1.92 & 7.53 & 1.19 \\ \cline{1-5} 
2 & 0.23$\pm$0.22 & 2.88 & 8.47 & 1.45\\ \hline
\end{tabular}
\end{center}
\end{table}

This section describes the experiments we performed to determine the impact of each model component on the overall model performance. All experiments were conducted by training the model for 400 epochs. The synthetic data generated for patient 1 were used; hence all results should be compared with patient 1 values in Table \ref{table:summary_stat_table}). 

We first assessed the impact of the four MLPs in the image encoder. As described, our incorporated an MLP for each convolutional layer, with each MLP responsible for five features per vertex, for a total of 20 features. Two experiments were conducted: 1) we removed MLPs associated with the first two convolutional layers, resulting in only ten image features per vertex; 2) we removed all MLPs except the final one, associated with the final convolutional layer, resulting in five features per vertex. In each of these cases, projection angle information was integrated into the input layer. Results are shown in Table \ref{table:impact_of_MLPs}, which demonstrate the importance of all four MLPs. 

\begin{table}[ht]
\begin{center}
\caption{Impact of removing gantry (projection angle) information from the image encoder. All values reported in mm.}
\label{table:impact_of_gantry_info}
\begin{tabular}{|c|c|c|c|}
\hline
\textbf{Mean (std)} & \textbf{Mean peak} & \textbf{Max peak} & \textbf{$99^{\rm th}$ Percentile}\\
\hline
0.23$\pm$0.21 & 2.81 & 9.33 & 1.27 \\ \hline
\end{tabular}
\end{center}
\end{table}

We next explored the impact of projection angle information in the image encoder by omitting this information from the input layer. For this experiment, the number (i.e. four) of MLPs remained constant. The results in Table \ref{table:impact_of_gantry_info} show that feeding angle information to the input layer is more successful. 

\begin{table}[ht]
\begin{center}
\caption{Impact of removing skip connections from the mesh deformation network. All values reported in mm.}
\label{table:impact_of_residual_connections}
\begin{tabular}{|c|c|c|c|}
\hline
\textbf{Mean (std)} & \textbf{Mean peak} & \textbf{Max peak error} & \textbf{$99^{\rm th}$ Percentile}\\
\hline
0.20$\pm$0.19 & 2.04 & 7.69 & 1.31 \\ \hline
\end{tabular}
\end{center}
\end{table}

In the next experiment, we focused on the effect of residual connections used in the mesh deformation network by simply removing them. Results in Table \ref{table:impact_of_residual_connections} indicate our approach with residual connections was superior.

\begin{table}[ht]
\begin{center}
\caption{Impact of replacing graph-attention layers with graph convolutional network layers. All values reported in mm.}
\label{table:impact_of_graph_convs}
\begin{tabular}{|c|c|c|c|}
\hline
\textbf{Mean (std)} & \textbf{Mean peak} & \textbf{Max peak} & \textbf{$99^{\rm th}$ Percentile}\\
\hline
0.18$\pm$0.16 & 1.51 & 7.05 & 0.83 \\ \hline
\end{tabular}
\end{center}
\end{table}

Finally, we explored the impact of the graph attention layers in the mesh deformation network by replacing them with GCN layers. The architecture was otherwise unchanged. Results in Table \ref{table:impact_of_graph_convs} demonstrate that utilizing GAT layers is more effective.

\typeout{}
\bibliographystyle{unsrt}
\bibliography{DMMpaper.bib}

\end{document}